\documentclass[a4paper]{article}

\usepackage[a4paper, margin=1in]{geometry}
\usepackage{graphicx}

\usepackage{csquotes}
\usepackage[english]{babel}

\usepackage{amsmath,amsfonts}

\usepackage{authblk}
\usepackage{xcolor}
\usepackage{todonotes}
\usepackage{siunitx}
\usepackage{booktabs}

\usepackage[style=numeric,backend=biber,maxbibnames=10,minbibnames=3,
terseinits=true,abbreviate=true,
natbib=true,giveninits=true,autolang=hyphen,isbn=false]{biblatex}
\newcommand{\vx}{\boldsymbol{x}}
\newcommand{\vn}{\boldsymbol{n}}
\newcommand{\ve}{\boldsymbol{e}}
\newcommand{\vz}{\boldsymbol{z}}

\newcommand{\vtheta}{\boldsymbol{\theta}}

\newcommand{\vf}{\boldsymbol{f}}
\newcommand{\dd}{\mathrm{d}}
\newcommand{\Vm}{V_\mathrm{m}}
\newcommand{\phie}{\phi_\mathrm{e}}
\newcommand{\tG}{\mathbf{G}}
\newcommand{\tGe}{\tG_\mathrm{e}}
\newcommand{\tGi}{\tG_\mathrm{i}}
\DeclareMathOperator{\diverg}{div}

\newcommand{\NN}{\mathcal{N\!N}}
\newcommand{\NNlf}{\NN_\text{LF}}
\newcommand{\NNsdf}{\NN_\text{SDF}}

\begin{document}


\title{Learning geometry-dependent lead-field operators for forward ECG modeling}

\author[1]{Arsenii Dokuchaev\thanks{Corresponding author: arsenii.dokuchaev@unitn.it}}
\author[2]{Francesca Bonizzoni}
\author[2]{Stefano Pagani}
\author[2]{Francesco Regazzoni}
\author[1,3]{Simone Pezzuto}

\affil[1]{Laboratory of Mathematics for Biology and Medicine, Università di Trento, Italy}
\affil[2]{MOX Laboratory, Department of Mathematics, Politecnico di Milano, Italy}
\affil[3]{Euler Institute, Università della Svizzera italiana, Switzerland}
\date{}

\maketitle

\begin{abstract}

Modern forward electrocardiogram (ECG) computational models rely on an accurate representation of the torso domain. The lead-field method enables fast ECG simulations while preserving full geometric fidelity. Achieving high anatomical accuracy in torso representation is, however, challenging in clinical practice, as imaging protocols are typically focused on the heart and often do not include the entire torso. In addition, the computational cost of the lead-field method scales linearly with the number of electrodes, limiting its applicability in high-density recording settings. To date, no existing approach simultaneously achieves high anatomical fidelity, low data requirements and computational efficiency.
In this work, we propose a shape-informed surrogate model of the lead-field operator that serves as a drop-in replacement for the full-order model in forward ECG simulations. The proposed framework consists of two components: a geometry-encoding module that maps anatomical shapes into a low-dimensional latent space, and a geometry-conditioned neural surrogate that predicts lead-field gradients from spatial coordinates, electrode positions and latent codes.
The proposed method achieves high accuracy in approximating lead fields both within the torso (mean angular error $\SI{<5}{\degree}$) and inside the heart, resulting in highly accurate ECG simulations (relative mean squared error $\SI{<2.5}{\percent}$). The surrogate consistently outperforms the widely used pseudo lead-field approximation while preserving negligible inference cost. Owing to its compact latent representation, the method does not require a fully detailed torso segmentation and can therefore be deployed in data-limited settings while preserving high-fidelity ECG simulations.

\end{abstract}

\section{Introduction}

In cardiology, electrocardiography records electric potentials originating from the heart over the body surface. A classic example is the 12-lead surface electrocardiogram (ECG or EKG), in which potentials are recorded at standardized chest locations, comprising ten electrode positions, including the ground (reference) electrode. Body Surface Potential Maps (BSPMs) are a spatially denser version of ECG measurements and are typically employed to solve the inverse problem of electrocardiography, which enables non-invasive reconstruction of cardiac potentials or sources~\cite{franzone1978approach, ramanathan2004noninvasive, li2024solving}.

From a modeling perspective, the translation of cardiac sources into an ECG or BSPM is typically achieved using the lead-field method~\cite{Potse2018-iu}. Lead fields form the basis of efficient forward-modeling workflows in cardiac electrophysiology as well as of the associated inverse problem~\cite{Pezzuto2017-yw, ECGTwin2025}. Mathematically, the lead field for a single lead corresponds to the solution of the adjoint pseudo-bidomain problem associated with a unit current injection at the electrode pair and is closely related to the Green's function of the pseudo-bidomain operator. Therefore, it strongly depends on heart shape, heart orientation and position within the torso, electrode locations, and overall anatomy around the heart~\cite{Zappon2025-nm}.

Setting up and solving the lead-field problem is, however, non-trivial for several reasons. First, in standard cardiac imaging, high-quality clinical images for torso segmentation and anatomy preparation are often incomplete or missing, because imaging primarily targets the heart. This issue can be partially circumvented by ``implanting'' patient-specific cardiac anatomy into a template torso model, for instance from a previously segmented case~\cite{VicentePuig2026, qian2025developing}, or by using statistical shape models conditioned on available data~\cite{li2025personalized}. Notably, electrode positions must also be segmented to be included in the model; however, electrodes are generally removed during imaging, so precise localization is often difficult~\cite{bergquist2023uncertainty}.

Second, the overall assembly cost of the transfer operator is computationally non-negligible and scales linearly with the number of electrodes (or, more precisely, the number of leads). Computing the lead-field transfer operator requires solving an elliptic partial differential equation (PDE) over the whole-body domain for each independent lead configuration, accounting for electrical conductivities (possibly anisotropic), the shapes of major organs and cavities, and electrode locations~\cite{Potse2018-iu}. The lead-field problem therefore requires careful segmentation of the entire torso (heart, lungs, major blood cavities, liver, fat, ribs) for electrical conductivities~\cite{keller2010ranking} and accurate placement of electrodes on the torso surface.

A typical strategy to circumvent both issues is the use of approximate anatomical models, which relax imaging requirements. The pseudo lead-field approach only requires electrode positions relative to the heart and has a closed-form solution that is very fast to compute~\cite{camps2025harnessing}.
However, torso anatomy is known to affect lead fields (and therefore simulated ECGs), especially for electrodes close to the heart, such as precordial leads. Intermediate approaches exist and are based on surface meshes of the body, heart, and possibly lungs. They usually employ the boundary element method, which does not require meshing the three-dimensional volume. However, this approach cannot be trivially extended to support anisotropy and may suffer from numerical instabilities due to the presence of singular kernels in the boundary integral formulation and the resulting ill-conditioning of the system matrices~\cite{wang2006application}. Therefore, existing approach cannot guarantee high anatomical fidelity while maintaining low data requirements and computational efficiency.

In this work, we propose a methodology to approximate the lead-field operator efficiently while retaining high accuracy for practical ECG applications. Importantly, the proposed approach does not rely on a detailed volumetric torso segmentation, but instead leverages a compact geometric representation that remains applicable in limited-data settings. 
Our approach is inspired by neural field methods (or implicit neural representation), which parameterize continuous physical fields defined over general $N$-dimensional domains using neural networks. Such methods have been widely adopted in 2-D and 3-D image synthesis, reconstruction and rendering tasks~\cite{Xie2022-vn}. For example, DeepSDF learns a continuous signed distance function (SDF) representation of shapes, enabling reconstruction, interpolation, and completion from partial or noisy 3-D data \cite{Park2019-nj}. Similar neural field approaches have proven effective for the parameterization and reconstruction of cardiac geometries~\cite{sander2023reconstruction}.
Specifically, our method has two components: one for geometric encoding and one for lead-field prediction. The first is a DeepSDF model~\cite{Park2019-nj, verhulsdonk24a}, based on an auto-decoder neural network that predicts the Signed Distance Function (SDF) of a shape for a given latent code. During training, the network learns a joint latent representation of torso and heart shapes. We also consider, for comparison, a more standard geometric representation based on Principal Component Analysis (PCA), where the latent representation is the linear subspace spanned by the principal components. 
The second component is a neural implicit representation of the lead-field function, conditioned on the geometry latent code (either DeepSDF- or PCA-based) and electrode coordinates~\cite{nagel2021bi, rodero2021linking}. The approach is fully data-driven in both components and does not explicitly encode physical constraints in the model architecture or loss formulation. Training requires only a dataset of precomputed lead-field solutions, which are here generated using a high-fidelity torso model for multiple heart-torso configurations. Finally, we compare ECGs computed with this approach against ECGs based on the pseudo lead field.

\section{Methods}

\begin{figure}[htbp]
    \centering
    \includegraphics[width=0.8\textwidth]{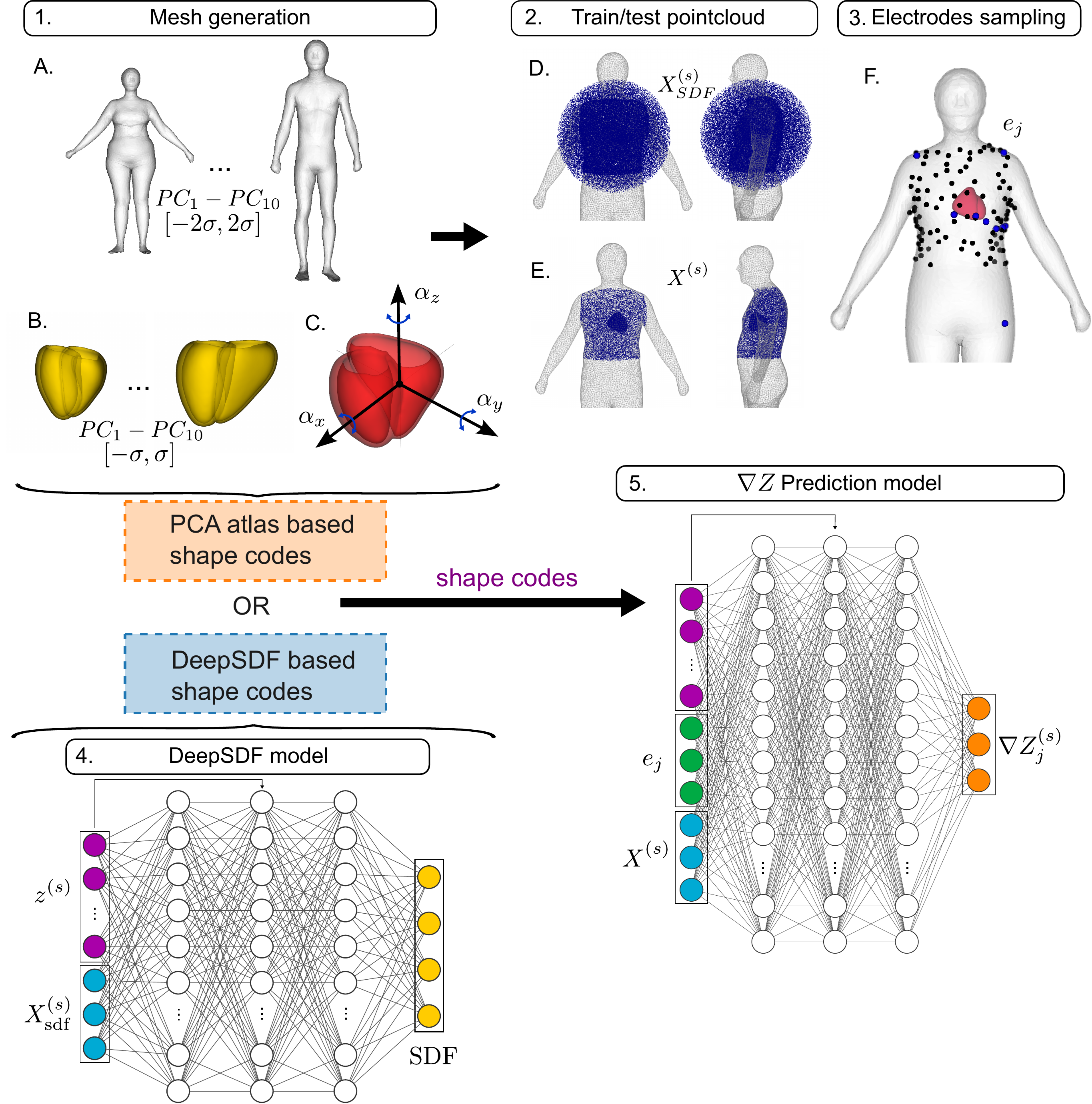}
    \caption{
    Graphical representation of the developed pipeline.
    (1.) Mesh generation stage: we used torso and heart models from a statistical PCA atlas. To create joint models, we varied (A.) the first 10 principal components of the torso, (B.) the first 10 principal components of the heart, and (C.) heart rotation angles along anatomical axes: $\alpha_x$ along the LV--RV axis, $\alpha_y$ along the anterior-posterior axis, and $\alpha_z$ along the LV long axis.  As a result, a set of geometries was obtained and used to generate training and test point clouds (2.), while a set of features (10 torso modes + 10 heart modes + 3 angles) was used as feature vectors describing each dataset (highlighted with an orange frame).
    (2.) Train/test point-cloud generation. (D.) Point cloud around and inside the torso used to train/test the DeepSDF model. (E.) Point cloud inside the torso used to train/test the $\nabla Z$ prediction model.
    (3.) Variation of unipolar electrode location on the anterior surface of the torso.
    (4.) Schematic representation of DeepSDF model used for generation of DeepSDF based shape codes (highlighted with a blue frame).
    (5.) Schematic representation of lead field gradient $\nabla Z$ prediction model.
    }
    \label{fig:lf_architecture}
\end{figure}

This section presents the mathematical and computational methods used for preprocessing geometric models, generating training and test samples, and designing and training the proposed neural network architectures.
We begin by introducing the mathematical formulation of the lead-field problem and the corresponding forward ECG computation. We then describe the geometric pipeline, including the generation of joint heart-torso anatomies and the sampling of training and test point clouds. Next, we detail the numerical computation of ground-truth lead fields via finite element approximation of the forward problem and the architecture and training of the proposed neural networks. Finally, we provide a description of the pseudo lead-field formulation used for comparison.

An overview of the complete pipeline is shown in Fig.~\ref{fig:lf_architecture}. The workflow consists of:
(1) generating joint heart-torso geometries and sampling training and test point clouds;
(2) defining electrode configurations;
(3) computing reference lead fields by solving the forward problem using the finite element method (FEM);
(4) training a neural network to learn a compact anatomical representation via DeepSDF; and
(5) training a second neural network to predict lead-field gradients from spatial coordinates, electrode positions, and the learned latent codes.

The surrogate model is subsequently in-silico validated by comparing predicted lead-field gradients and the resulting ECGs against FEM-based reference solutions for previously unseen geometries and activation patterns.


\subsection{Forward ECG modeling and lead-field problem}

%

Let denote by $\Omega \subset \mathbb{R}^3$ the body domain and by $\Sigma = \partial\Omega$ the body surface; the active myocardium of the heart is $\Omega_H \subset \Omega$, while the passive conductor tissue is $\Omega_0 := \Omega\setminus \bar{\Omega}_H$ and $\Gamma = \partial\Omega_H$ is the heart-torso surface. Given the transmembrane potential $\Vm(\vx,t)$ on $\Omega_H$, the extracellular potential $\phi_0(\vx,t)$ and $\phi_e(\vx,t)$, respectively in the passive and active tissue, solves the following pseudo-bidomain equation:
\begin{equation}
\label{eq:pseudo}
\left\{ \begin{aligned}
- \diverg\bigl(\tG \nabla \phie \bigr) &= \diverg(\tGi \nabla\Vm), && \vx \in \Omega_H, t \in \mathbb{R}, \\
- \diverg\bigl(\tG_0 \nabla \phi_0 \bigr) &= 0, && \vx \in \Omega_0, t \in \mathbb{R}, \\
\phie - \phi_0 &= 0, && \vx \in \Gamma, t \in \mathbb{R}, \\
\tG_0 \nabla\phi_0 \cdot\vn - \tG \nabla \phie\cdot\vn &= \tGi\nabla \Vm \cdot\vn, && \vx \in \Gamma, t \in \mathbb{R}, \\
\tG_0 \nabla\phi_0 \cdot\vn &= 0, && \vx \in \Sigma, t \in \mathbb{R},
\end{aligned} \right.
\end{equation}
where $\tGi$, $\tG$ and $\tG_0$ are respectively the intra-cellular, bulk and torso conductivity tensors, and $\vn$ is the outwards normal with respect to the heart-torso surface. In general, the bulk conductivity is $\tG = \tGi+\tGe$, where $\tGe$ is the extra-cellular conductivity. The conductivity tensors encode the anisotropy of the tissue, especially for the myocardium. In what follows, we are assuming:
$$
\begin{aligned}
\tGi &= \sigma_{i,t}\mathbf{I} + (\sigma_{i,f} - \sigma_{i,t}) \vf\otimes\vf, \\
\tGe &= \sigma_{e,t}\mathbf{I} + (\sigma_{e,f} - \sigma_{e,t}) \vf\otimes\vf, \\
\tG_0 &= \sigma_0 \mathbf{I}, \\
\end{aligned}
$$
with $\vf(\vx)$ the local fiber direction in the myocardium, and $\sigma_{i,t}$, $\sigma_{e,t}$, $\sigma_{i,f}$ and $\sigma_{e,f}$ respectively the intra- and extra-cellular fiber and transverse electric conductivities, and $\sigma_0$ the torso conductivity. The values used in our simulations are based on \cite{Potse2006-wf}, specifically:
$\sigma_{i,t}=\SI{0.3}{\milli\siemens\per\cm}$,
$\sigma_{i,f}=\sigma_{e,f}=\SI{3.0}{\milli\siemens\per\cm}$ and
$\sigma_{e,t}=\SI{1.2}{\milli\siemens\per\cm}$.
For the torso we considered $\sigma_0 = \SI{0.6}{\milli\siemens\per\cm}$, according to \cite{Boulakia2010-vz}.

The solution to the system \eqref{eq:pseudo} is defined up to a constant, which can be fixed by setting a reference node or, equivalently, by constraining the solution to have zero average on the body surface:
\begin{equation} \label{eq:null}
\int_\Sigma \phi_0 \: \dd\vx = 0.
\end{equation}
The ECG is generally formed by a set of leads, each obtained from a linear combination of $\phi_0(\vx,t)$ measured at some electrodes locations $\ve_j \in \Sigma$. (Under suitable assumptions on the domain, coefficients and $\Vm$, it is possible to prove that the potential is globally $\mathcal{C}(\bar{\Omega})$, thus point-wise evaluation is well-defined. See~\cite[Thm.~5.1]{franzone_mathematical_2014}.) For each lead $\ell=1,\ldots,L$, the single lead ECG reads
$$
V_\ell(t) = \sum_{j=1}^{n_\ell} \alpha_j \phi_0(\ve_j, t),
$$
where $n_\ell$ is number of electrodes for lead $\ell$, and $\alpha_k$ are the zero-sum weights. Note that the evaluation of the ECG requires the numerical solution of \eqref{eq:pseudo} for each time step  of the transmembrane potential.

\begin{figure}[htbp]
    \centering
    \includegraphics[width=0.5\linewidth]{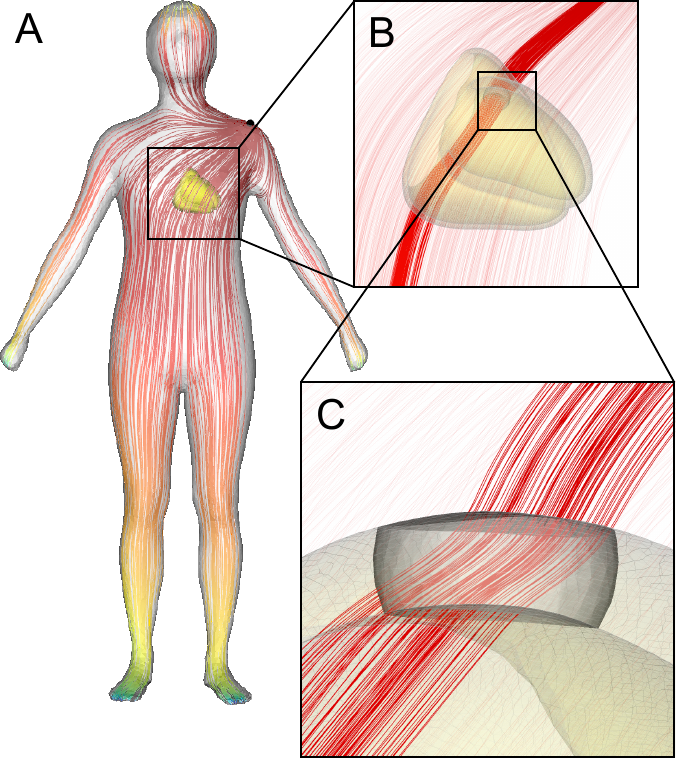}
    \caption{Computed lead field for a unipolar lead (left shoulder), visualized using streamlines representing the direction of $\nabla Z$.
    (A.) Streamlines shown throughout the entire torso. 
    (B.) Close-up view of the heart region (the heart is shown in yellow); streamlines passing through the free wall of the left ventricle (LV) are highlighted in bold red.
    (C.) Enlarged view of the LV free wall (represented as a cylinder) with the corresponding gradient streamlines highlighted in bold. Note the change in the slope of the streamlines at the heart–torso interface, reflecting the change in direction of $\nabla Z$ across the boundary.}
    \label{fig:leadfield_streamlines}
\end{figure}

The lead field method is a natural way to alleviate the computational cost of evaluating the ECG~\cite{Geselowitz1989-vw, Pezzuto2017-yw, Potse2018-iu}. In fact, it is possible to prove that the solution with zero reference potential of \eqref{eq:pseudo} is given by~\cite[Prop.~5.1]{franzone_mathematical_2014}
\begin{equation}
\label{eq:ECG}
\phi_0(\ve_j, t) = - \int_{\Omega_H} \tGi \nabla \Vm(\vx,t) \cdot \nabla Z(\vx; \ve_j)\: \dd\vx,
\end{equation}
where $Z(\vx; \ve_j)$ is the \emph{lead-field function} with respect to the electrode $\ve_j$, and solves the problem:
\begin{equation}
\label{eq:leadfield}
\left\{ \begin{aligned}
- \diverg\bigl(\tG \nabla Z \bigr) &= 0, && \vx \in \Omega_H, \\
- \diverg\bigl(\tG_0 \nabla Z_0 \bigr) &= 0, && \vx \in \Omega_0, \\
Z - Z_0 &= 0, && \vx \in \Gamma, \\
\tG_0 \nabla Z_0 \cdot\vn - \tG \nabla Z \cdot\vn &= 0, && \vx \in \Gamma, \\
\tG_0 \nabla Z_0 \cdot\vn &= \delta_{\ve_j}(\vx) - |\Sigma|^{-1}, && \vx \in \Sigma,
\end{aligned} \right.
\end{equation}
where $\delta_{\ve_j}(\vx)$ is the Dirac's delta centered at $\ve_j$. Note that this problem does not depend neither on time nor the transmembrane potential. An example of the (gradient of the) lead-field function computed from a single electrode is provided in Fig.~\ref{fig:leadfield_streamlines}.
With a slight abuse of notation, we will denote simply by $Z(\vx)$ the lead field in the whole body, without making a distinction between $Z(\vx)$ and $Z_0(\vx)$.

\subsection{Data preparation}


We used the bi-ventricular statistical shape model (SSM) proposed in \cite{Bai2015-fb} as the baseline heart model. 
This statistical atlas was constructed from MR images of \num{1093} healthy subjects and provides 100 principal components with associated variances describing the joint distribution of the left ventricular myocardium (LV) and right ventricular blood pool (RV) surfaces. 
While the original atlas contained separate surface representations, we adopted a modified version described in \cite{schuler_2021_4506463}.

\begin{figure}[htbp]
    \centering
    \includegraphics[width=0.9\linewidth]{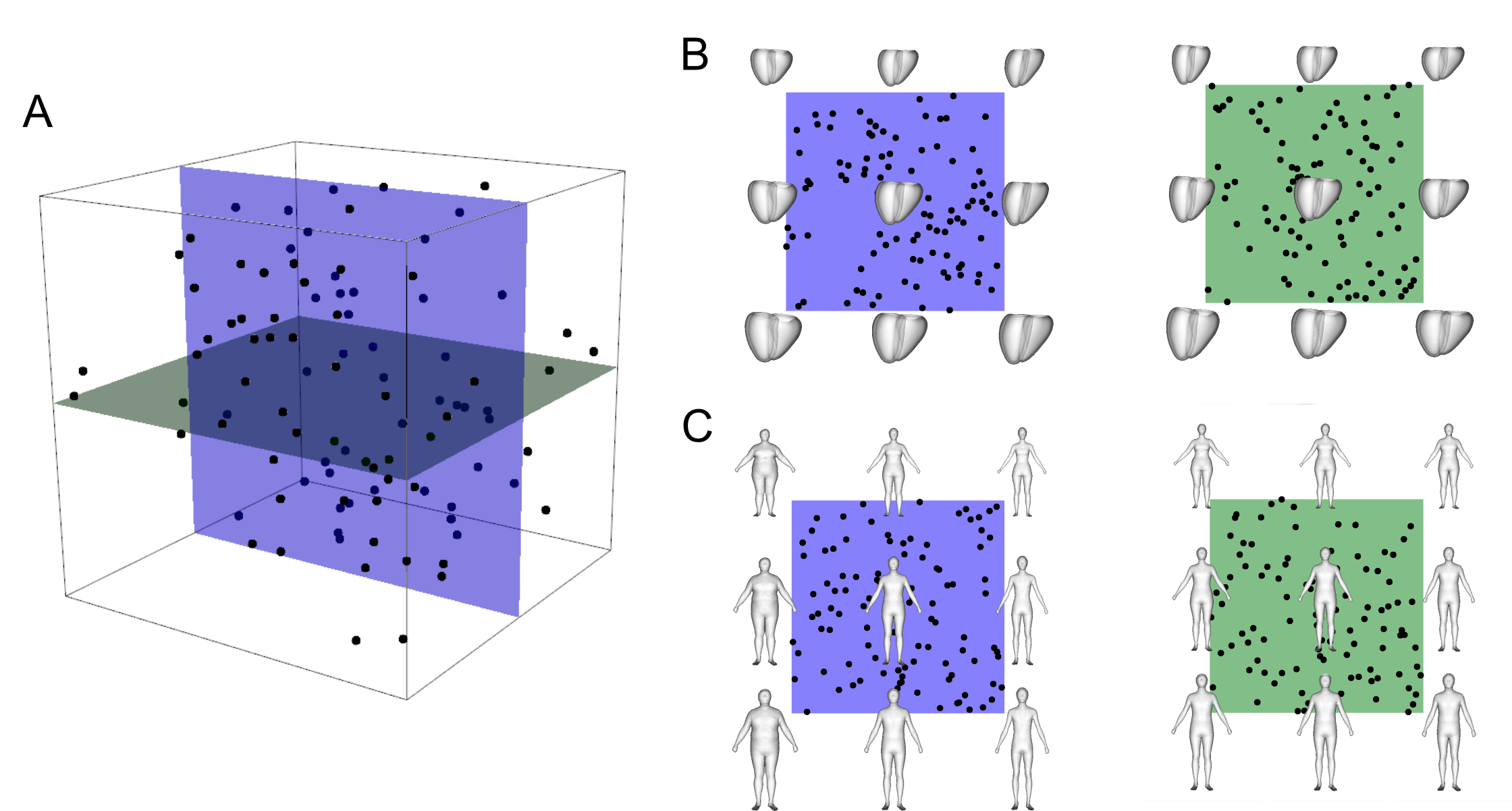}
    \caption{Design of training samples for parametric models of the torso and heart. \textbf{A} Projection of a 10-dimensional parametric space onto a 3D cube, obtained by neglecting the last 7 dimensions. Black dots indicate the position of 100 sampling points selected using Latin hypercube sampling from a uniform distribution. \textbf{B} and \textbf{C} are cross-sections of the parametric space for the heart (\textbf{B}) and torso (\textbf{C}) showing the variations of the three principal components: PC1, PC2 in blue colours, PC1, PC3 in green colours.}
    \label{fig:pcaplot}
\end{figure}

To construct a dataset of 100 heart geometries, we varied the weights of the first 10 principal components within the range $[-1, 1]$ standard deviations using Latin hypercube sampling with a uniform distribution. 
A uniform distribution was chosen instead of a Gaussian one to ensure broad and even coverage of the shape space, including geometries located near the boundaries of the selected range. 
Latin hypercube sampling guarantees uniform coverage of the resulting 10-dimensional parameter space (see Fig.~\ref{fig:pcaplot}).

For each heart geometry, myocardial fiber orientations were assigned using the rule-based method of \cite{Bayer2012-os}, with fiber angles $\alpha_{\text{endo}} = 40^\circ$, $\alpha_{\text{epi}} = -50^\circ$, and sheet angles $\beta_{\text{endo}} = -65^\circ$, $\beta_{\text{epi}} = 25^\circ$.

Because the lead field depends not only on electrode configuration but also on the geometry of the computational domain \cite{Potse2018-iu}, we also varied torso shape. 
Torso geometries were generated from the MPII Human Shape statistical atlas \cite{pishchulin17pr}, using a sampling strategy analogous to that employed for the heart.

Even among healthy subjects, the size, position, and orientation of the heart significantly influence the ECG, as demonstrated in clinical studies \cite{Hoekema1999-zh} and computational investigations \cite{Nguyen2015-jn, Minchole2019-fi, Zappon2025-nm}. 
To account for this variability, we additionally varied heart translation and rotation within the torso.
The baseline heart position was defined by aligning the heart centroid with a reference point inside the torso volume. 
The baseline orientation was specified by three rotation angles $\alpha_X$, $\alpha_Y$, and $\alpha_Z$ describing rotations of the cardiac long axis relative to the anatomical planes, following \cite{Odille2017-qm} (see Fig.~\ref{fig:lf_architecture}C).

Three-dimensional finite element meshes were generated using Gmsh \cite{Geuzaine2009-za}. 
The characteristic element sizes were \SI{0.8}{\mm} for the heart and \SI{11.8}{\mm} for the torso. 
The use of statistical atlases significantly simplified and automated geometry processing, including the identification of subendocardial surfaces, base and apex regions, and other anatomical landmarks required to compute Laplace-Dirichlet scalar fields for myocardial fiber assignment \cite{Bayer2012-os}. 
In addition, atlas-based modeling enabled standardized placement of electrodes on the torso surface.
On the anterior torso surface of each mesh, 100 unipolar electrodes were uniformly sampled. 
In addition, the nine independent electrodes of a standard 12-lead ECG configuration were placed for each geometry (see Fig.~\ref{fig:lf_architecture}, block~3).

We employed a fixed train-test split at the level of virtual patients, ensuring that no samples from a test geometry were used during training.

\subsection{Lead-field neural surrogate}

Since the ECG signal $V(t)$ depends on the lead-field gradient $\nabla Z$ within the heart region $\Omega_H$, we focus on parameterizing and reconstructing the gradient field $\nabla Z(\mathbf{x})$ throughout the domain. Note that in general $\nabla Z$ has a jump discontinuity on the heart-torso interface, due to the different conductivity of the heart and its surrounding tissue.

The problem of learning the lead-field gradient conditioned on anatomical variability can be formulated as approximating a conditional vector-valued function
$$
\NNlf \colon \mathbb{R}^3 \times \mathbb{R}^3 \times \mathbb{R}^{d_z} \rightarrow \mathbb{R}^3,
$$
which maps a spatial point $\vx \in \Omega^{(p)}$, an electrode position $\ve_j \in \Sigma^{(p)}$, and a latent anatomical code $\vz^{(p)} \in \mathbb{R}^{d_z}$ associated with geometry $(p)$ to the corresponding lead-field gradient:
\[
\NNlf \bigl(\vx, \ve_j, \vz^{(p)}; \vtheta_\text{LF} \bigr) \approx \nabla Z_j^{(p)}(\vx),
\]
where $Z_j^{(p)}$ is the lead field (with respect to electrode $\ve_j$) for the domain $\Omega^{(p)}$. The latent code will encode for the domain $\Omega^{(p)}$, which includes the heart and other organs, and the electric conductivities, including the fiber direction. The weights of the neural networks are denoted by $\vtheta_\text{LF}$.

We model $\NNlf$ using a fully connected neural network (see Fig.~\ref{fig:lf_architecture}, block~5). For each virtual patient $(p)$, we randomly sampled $2^{16}$ spatial points $\vx \in \Omega^{(p)}$ within the torso domain and computed the corresponding gradients $\nabla Z_j^{(p)}(\vx)$ for each electrode $\ve_j^{(p)}$. Sampling was biased toward the torso surface and the heart-torso interface (Fig.~\ref{fig:leadfield_streamlines}), where the gradient exhibits sharp spatial variations: approximately \SI{80}{\percent} of the sampled points lie within \SI{10}{\mm} of these interfaces. The set of sampled points $\mathbf{X}^{(p)}$ differs across patients but is shared among all electrodes of a given patient.


To encode the joint heart-torso geometry, we investigated two strategies:
\begin{itemize}
    \item A \textbf{PCA-based encoding}, with latent anatomical code as:
    $$
    \vz^{(p)} = \begin{bmatrix}
      w^\text{heart}_1, \ldots, w^\text{heart}_{N_h},
      w^\text{torso}_1, \ldots, w^\text{torso}_{N_t},
      \alpha_x, \alpha_y, \alpha_z
    \end{bmatrix}^\text{T},
    $$
    where $\{ w^\text{heart}_i \}$ and $\{ w^\text{torso}_i \}$ are the principal component (PC) weights for the heart and torso shapes, and $\alpha_x$ (left-right), $\alpha_y$ (anterior-posterior), and $\alpha_z$ (apex-base) are the heart rotation angles along standard anatomical axes (see Fig.~\ref{fig:lf_architecture}, block~1).
    \item A \textbf{DeepSDF-based encoding}, with latent anatomical code corresponding to the learned shape codes obtained from the DeepSDF model (see Sec.~\ref{sec:DeepSDF}).
\end{itemize}

In addition to spatial coordinates and latent codes, the electrode position $\ve_j^{(p)}$ was included as part of the network input. To enhance generalization across different anatomies, electrode locations were expressed in a normalized torso coordinate system $\tilde \ve_j = (x_j, y_j, z_j)$, where $x_j \in [-1,1]$ (left-right direction), $y_j \in [0,1]$ (posterior-anterior direction), and $z_j \in [0,1]$ (superior-inferior direction).

Architecturally, the neural network comprises five hidden layers with 256 neurons each and ReLU activation functions. In the third hidden layer, the latent code and electrode coordinates are concatenated to the intermediate feature representation. The final layer is linear with three outputs corresponding to the $x$, $y$ and $z$ components of the gradient vector.

Training was performed using the ADAM optimizer \cite{kingma2014adam} to minimize the loss function
\begin{equation}
\mathcal{L}(\vtheta_\text{LF}) =
\frac{1}{N}\frac{1}{N_{\text{ele}}}
\sum_{p=1}^{N}
\sum_{j=1}^{N_{\text{ele}}}
\mathbb{E}_{\vx \sim \mathbf{X}^{(p)}}
\Big[
\mathcal{L}_\text{MSE}(\vx,\ve_j,\vz^{(p)}, \vtheta_\text{SDF}) +
\lambda_{\text{cos}} \mathcal{L}_{\text{cos}}\bigl(
f_\theta(\vx, \ve_j^{(p)}, \vz^{(p)}), \nabla Z_j^{(p)}(\vx)\bigr)
\Big],
\end{equation}
where
$$
\mathcal{L}_\text{MSE} = 
\bigl\| \NNlf(\vx, \ve_j, \vz^{(p)}; \vtheta_\text{SDF}) - \nabla Z_j^{(p)}(\vx) \bigr\|_2^2
$$
and $\mathcal{L}_{\text{cos}}$ is a cosine similarity loss, introduced to improve the fitting of the gradient direction when its magnitude is small (and thus not captured by the first term). The cosine similarity is defined for vectors $\hat{\mathbf{v}}$ and $\mathbf{v}$ as
\[
\mathcal{L}_{\text{cos}}(\hat{\mathbf{v}}, \mathbf{v}) =
1 -
\frac{\hat{\mathbf{v}} \cdot \mathbf{v}}
{\|\hat{\mathbf{v}}\|_2 \|\mathbf{v}\|_2 + \varepsilon}.
\]
To better capture high-frequency spatial variations,  particularly near the heart-torso interface, we additionally employed Fourier features \cite{Tancik2020-am} as a positional encoding mechanism.

\subsection{PCA-based geometry encoding}

As a low-dimensional representation of the joint heart–torso geometries, we used the same parameter vectors that were employed to generate the shapes. 
Specifically, these include the weights of the first 10 principal components of the torso, the weights of the first 10 principal components of the heart geometry, and three rotation angles describing the orientation of the heart with respect to the anatomical axes.

The resulting 23-dimensional vector was used as the latent code and provided as input to the neural surrogate $\NNlf$, together with the spatial coordinates of the evaluation points and the electrode positions.

\subsection{DeepSDF geometry encoding}
\label{sec:DeepSDF}

To represent the joint anatomical geometries of the torso and heart, we adapted the DeepSDF model with an auto-decoder architecture \cite{Park2019-nj, verhulsdonk24a}. 
This framework encodes complex multi-object geometries into a continuous low-dimensional latent representation. 
The decoder network is trained to approximate signed distance functions (SDFs) corresponding to four anatomical surfaces: the torso, the left ventricular endocardium (LV), the right ventricular endocardium (RV), and the epicardium.

We denote the DeepSDF decoder by
\[
\NNsdf \colon \mathbb{R}^3 \times \mathbb{R}^{d_z} \rightarrow \mathbb{R}^4,
\]
where $\vx \in \mathbb{R}^3$ is a spatial coordinate and $\vz \in \mathbb{R}^{d_z}$ is a latent code describing the joint heart-torso geometry. The weights of the network are denoted by $\vtheta_\text{SDF}$.
The network outputs four signed distance functions, one for each surface, evaluated at location $\vx$ for the geometry encoded by $\vz$.

The model consists of five fully connected layers with 256 neurons each. 
The input is the concatenation of the latent code $\mathbf{z}$ (dimension $d_z = 16$) and the spatial coordinate $\mathbf{x}$. 
To preserve spatial information across layers, the coordinate vector is additionally concatenated with the output of the third hidden layer (skip connection). 
Fourier feature encoding \cite{Tancik2020-am} is applied to improve the representation of high-frequency geometric details, and Lipschitz layers with Lipschitz regularization \cite{Liu2022-zk} are used to enhance stability and smoothness of the learned latent space. 
ReLU activations and the Adam optimizer are employed during training.

For each geometry $(p)$, let
$$
\mathbf{X}_{\mathrm{SDF}}^{(p)} = \bigl\{ (\vx_n^{(p)}, s_n^{(p)}) \bigr\}_{n=1}^{N_p}
$$
denote sampled spatial points and their corresponding ground-truth SDF values. 
The training objective jointly optimizes decoder parameters $\phi$ and latent codes $\vz^{(p)}$:
\begin{multline}
\mathcal{L}\bigl(\vtheta_\text{SDF}, \{\vz^{(p)}\}\bigr) =
\frac{1}{N}
\frac{1}{N_p}
\sum_{p=1}^{N}
\sum_{n=1}^{N_p}
\left|
\NNsdf(\vx_n^{(p)}, \vz^{(p)}; \vtheta_\text{SDF}) - s_n^{(p)}(\vx_n^{(p)})
\right|^2 \\
+
\lambda_\text{prior} \frac{1}{N_p}
\sum_{n=1}^{N_p}\|\vz^{(p)}\|_2^2
+
\lambda_\text{Lip} \mathcal{L}_{\mathrm{Lip}}(\vtheta_\text{SDF}),
\end{multline}
where the first term corresponds to the mean squared reconstruction error between predicted and ground-truth SDF values.  The second term imposes a Gaussian prior on the latent codes, and $\mathcal{L}_{\mathrm{Lip}}$ denotes the Lipschitz regularization term weighted by $\lambda_\text{Lip}$.

\paragraph{Latent code inference.}
After training, the decoder parameters $\vtheta_\text{SDF}$ are kept fixed. 
Given a new joint geometry represented by sampled SDF values $\{(\vx_n, s_n)\}_{n=1}^{N}$ (e.g., derived from a point cloud or segmented MRI), the corresponding latent code is obtained by maximum a posteriori (MAP) estimation:
\begin{equation}
\label{eq:inference}
\vz^\star =
\operatornamewithlimits{arg\,min}_{\vz}
\left(
\frac{1}{N}
\sum_{n=1}^{N}
\left|
g_\phi(\vx_n, \vz) - s_n
\right|^2
+
\lambda_\text{Maha}
(\vz - \boldsymbol{\mu})^\top
\Sigma^{-1}
(\vz - \boldsymbol{\mu})
\right),
\end{equation}
where $\boldsymbol{\mu}$ and $\Sigma$ denote the empirical mean and covariance of the latent codes estimated from the latent codes obtained after then training phase, and $\gamma$ is the regularization parameter.
The Mahalanobis regularization term enforces anatomical plausibility by penalizing latent codes that are statistically distant from the learned distribution. 
This prior accounts for correlations between latent dimensions and improves robustness to noise in the input geometry.


\subsection{Evaluation metrics}
\label{sec:metrics}

For the DeepSDF network $\NNsdf$, we evaluated reconstruction quality using the Chamfer distance.
For each of the four predicted SDF fields (LV, RV, epicardium, and torso), both the predicted and ground-truth SDF values were thresholded by retaining only points with SDF $\leq 0$, corresponding to the interior of the surfaces.
The resulting point clouds were then compared using the Chamfer distance as a similarity metric.

Specifically, for two point clouds $X$ and $Y$ containing $N_X$ and $N_Y$ points, respectively, the Chamfer distance is defined as
\begin{equation}
\label{eq:cd}
\operatorname{CD}(X, Y) = \frac{1}{2} \left( \frac{1}{N_{X}} \sum_{x \in X} \min_{y \in Y} \| x - y \|_2 + \frac{1}{N_{Y}} \sum_{y \in Y} \min_{x \in X} \| y - x \|_2  \right).
\end{equation}

For the neural surrogate $\NNlf$, we assess surrogate performance at two complementary levels.

\begin{itemize}

\item \textbf{Lead-field (LF) metrics.}  
We quantify the accuracy of the predicted lead-field gradients by computing the mean angular error and the mean magnitude error over sampled spatial points.

For a given geometry $s$ and electrode $j$, the angular error at point $\vx \in \mathbf{X}^{(s)}$ is defined as
$$
\theta_j^{(s)}(\vx) =
\arccos
\left(
\frac{
\nabla \hat{Z}_j^{(s)}(\vx) \cdot \nabla Z_j^{(s)}(\vx)
}{
\|\nabla \hat{Z}_j^{(s)}(\vx)\|_2
\|\nabla Z_j^{(s)}(\vx)\|_2
+ \varepsilon
}
\right).
$$
The mean angular error is obtained by averaging $\theta_j^{(s)}(\vx)$ over all sampled points $\vx \in \mathbf{X}^{(s)}$ and over electrodes.
Magnitude errors are computed as the mean Euclidean difference
$$
\|\nabla \hat{Z}_j^{(s)}(\vx) - \nabla Z_j^{(s)}(\vx)\|_2,
$$
again averaged over spatial sampling points and electrodes.

We additionally report lead-field metrics computed on a restricted subset of electrodes, namely the six precordial leads of the standard 12-lead ECG. In this case, the averaging is performed only over this subset of electrodes.

\item \textbf{ECG-level metrics.}  
For each lead, we compare the predicted ECG waveform $\widehat V(t)$ with the FEM reference solution $V(t)$. 
We report the relative $\ell_2$ error over the time interval $[0,T]$,
$$
\frac{
\| \widehat V - V \|_{L^2(0,T)}
}{
\| V \|_{L^2(0,T)}
}.
$$
In addition, we evaluate differences in clinically relevant waveform characteristics, including QRS amplitude and QRS duration.
\end{itemize}

We note that the error in the lead-field gradient and the ECG error are directly related through Eq.~\eqref{eq:ECG}. 
Assume that both $V(t)$ and $\hat{V}(t)$ are computed from the same transmembrane potential $\Vm(\mathbf{x},t)$, but using two different lead-field functions, $Z(\mathbf{x})$ and $\hat{Z}(\mathbf{x})$, respectively. 
Then their difference satisfies
\begin{equation*}
V(t) - \hat{V}(t)
=
- \int_{\Omega_H}
\mathbf{G}_i \nabla \Vm(\mathbf{x},t)
\cdot
\nabla \big(Z(\mathbf{x}) - \hat{Z}(\mathbf{x})\big)
\, \mathrm{d}\mathbf{x}.
\end{equation*}
Applying the Cauchy-Schwarz inequality in $L^2(\Omega_H)$ yields
\begin{equation*}
|V(t) - \hat{V}(t)|^2
\le
\|\mathbf{G}_i \nabla \Vm(\cdot,t)\|_{L^2(\Omega_H)}^2
\,
\|\nabla (Z - \hat{Z})\|_{L^2(\Omega_H)}^2.
\end{equation*}
Integrating in time over $[0,T]$ gives
\begin{equation*}
\|V - \hat{V}\|_{L^2(0,T)}^2
\le
C \, \|\nabla (Z - \hat{Z})\|_{L^2(\Omega_H)}^2,
\end{equation*}
where the constant
$$
C_{\Vm}
=
\int_0^T
\|\tGi \nabla \Vm(\cdot,t)\|_{L^2(\Omega_H)}^2
\, \dd t
$$
depends only on the transmembrane potential and tissue conductivities.
Therefore, controlling the error in the lead-field gradient $\nabla Z$ directly yields a bound on the ECG error, through a multiplicative constant determined by the underlying electrophysiological activation.

\subsection{Experimental setup}
\label{sec:ECG}

The main goal of this work was to verify the feasibility and evaluate the quality of representing $\nabla Z(x)$ using a neural network, especially when employed to compute the ECG.

To model the transmembrane potential under different physiological and pathological conditions, we considered the following model:
$$
\Vm(\vx, t) = U\bigl(t - \tau(x)\bigr),
$$
where $U(\xi)$ is a template action potential, based on the ten Tusscher-Panfilov model of human ventricular cells~\cite{Ten_Tusscher2006-oc}, and $\tau(\vx)$ is the activation map satisfying the anisotropic eikonal equation~\cite{Colli_Franzone1990-hf, Franzone1993-kn}:
\begin{equation}
\left\{ \begin{aligned}
\sqrt{ \mathbf{V} \nabla\tau \cdot \nabla\tau } &= 1, && \vx \in \Omega_H, \\
\tau(\vx_i) &= \tau_i, && i = 1,\ldots,N_\text{stim},
\end{aligned} \right.
\end{equation}
with $\{ (\vx_i, t_i) \}$ being the initial set of activation sites and the symmetric positive definite tensor $\mathbf{V}(\vx)$ encoding for the anisotropic conduction velocity:
\begin{equation}
\mathbf{V}(\vx) = v_t^2\mathbf{I} + (v_f^2 - v_t^2) \vf \otimes \vf
\end{equation}
where $v_f$ and $v_t$ represents conduction velocities along the fiber direction and orthogonally to the fibers, respectively.

The eikonal model allows us to simulate different conditions depending on the boundary values $\{ \vx_i, t_i \}$. 
Two cases were studied:
(i) activation from two pacing sites, with the endocardium of the right ventricle near the apical region and the epicardium of the free wall of the left ventricle, mimicking the cardiac resynchronization therapy, and
(ii) simulation of sinus rhythm with the construction of a synthetic conduction system based on rule-based Purkinje network~\cite{Sahli_Costabal2016-su, Purkinje2025}.

We examined both the standard clinical electrode configuration for a 12-lead ECG and a denser set of unipolar electrodes distributed across the front of the torso, mimicking BSPMs and the possible uncertainty in the precordial electrodes of a 12-lead ECG.
Standard leads allow for direct comparison with clinical ECG morphology, while additional unipolar electrodes provide a wider choice of potential fields and are relevant for applications such as BSPMs and uncertainty in electrode placement.

The results were also compared with pseudo-lead-field, which is obtained from problem \eqref{eq:leadfield} assuming $\tG = \tG_0 = \sigma_0 \mathbf{I}$ and $\Omega = \mathbb{R}^3$ (infinite torso assumption). The exact solution reads as follows:
\begin{equation}
\label{eq:pseudoLF}
Z^{*}(\vx) = \frac{1}{4 \pi \sigma_0 \| \ve_j - \vx\|},
\end{equation}
Pseudo lead-field provides a simple approximation that requires no torso information, only the electrode locations.

\subsection{Implementation details}

All forward problems were solved on tetrahedral meshes using the finite element method implemented in the FEniCS framework~\cite{fenicsbook}. 
Since the lead-field formulation is defined up to an additive constant, the null space was removed by enforcing a reference potential constraint, see Eq.~\eqref{eq:null}. 
This ensures uniqueness of the numerical solution and stable computation of the gradient field.

The lead-field neural surrogate $\NNlf$ was trained for 800 epochs using the Adam optimizer with an initial learning rate of \num{1e-3}. After 400 epochs, the learning rate was reduced by a factor of two. 
The DeepSDF model $\NNsdf$ was trained for 2000 epochs using the same optimizer and initial learning rate of \num{1e-3}. No early stopping strategy was employed.
Training was performed on a single NVIDIA A100 GPU. 
The batch size was set to 10 for $\NNlf$ and 6 for the DeepSDF model. 
Due to the large number of sampled spatial points, training was computationally intensive: approximately 2 days were required for $\NNlf$ and 1.2 days for the DeepSDF model.
Finite element meshes contained approximately \num{40000} nodes and \num{150000} tetrahedra for the heart domain, and approximately \num{120000} nodes and \num{600000} tetrahedra for the full torso model.

\subsection{Software}

Finite element meshes were generated using Gmsh~\cite{Geuzaine2009-za}, while mesh preprocessing, postprocessing, and visualization were performed using PyVista~\cite{sullivan2019pyvista} and VTK~\cite{vtkBook}.
Surface remeshing was carried out using \texttt{pyacvd}.
Universal ventricular coordinates (UVCs), Laplace–Dirichlet rule-based (LDRB) myocardial fiber assignment, and lead-field computations were implemented in FEniCS. 
Eikonal simulations were performed using \texttt{fim-python}. 
Purkinje networks were generated using the \texttt{fractal-tree} package, and activation times at Purkinje–myocardial junctions (PMJs) were computed using \texttt{networkx}.
Signed distance functions and point cloud sampling were computed using the \texttt{mesh\_to\_sdf} library, which internally relies on \texttt{pyopengl}, \texttt{trimesh}, \texttt{pyrender}, and \texttt{pyDOE}. 
Latin hypercube sampling (LHS) of geometries and electrode positions was performed explicitly using \texttt{pyDOE}.
Both the DeepSDF model and the lead-field gradient surrogate were implemented in PyTorch~\cite{Ansel_PyTorch_2_Faster_2024}. Training was managed using PyTorch Lightning, with hyperparameter optimization performed via Optuna and experiment tracking conducted using MLflow.
Standard scientific computing tasks, including interpolation and nearest-neighbor search (KDTree) for Chamfer distance computation, were carried out using SciPy, NumPy, and scikit-learn. 
Visualization and plotting were performed using Vedo, PyVista, and Seaborn.


\section{Results}

\subsection{DeepSDF-based geometry encoding}
The DeepSDF model provides a compact latent representation of the joint heart-torso anatomy.
Reconstruction accuracy was evaluated as follows. Predictions from the DeepSDF decoder were computed on a uniform regular grid of $128^3$ spatial points. The predicted SDF values for the LV, RV, epicardium, and torso were then compared with the corresponding ground-truth SDF values interpolated onto the same $128^3$ grid.

\begin{figure}[htbp]
    \centering
    \includegraphics[width=\linewidth]{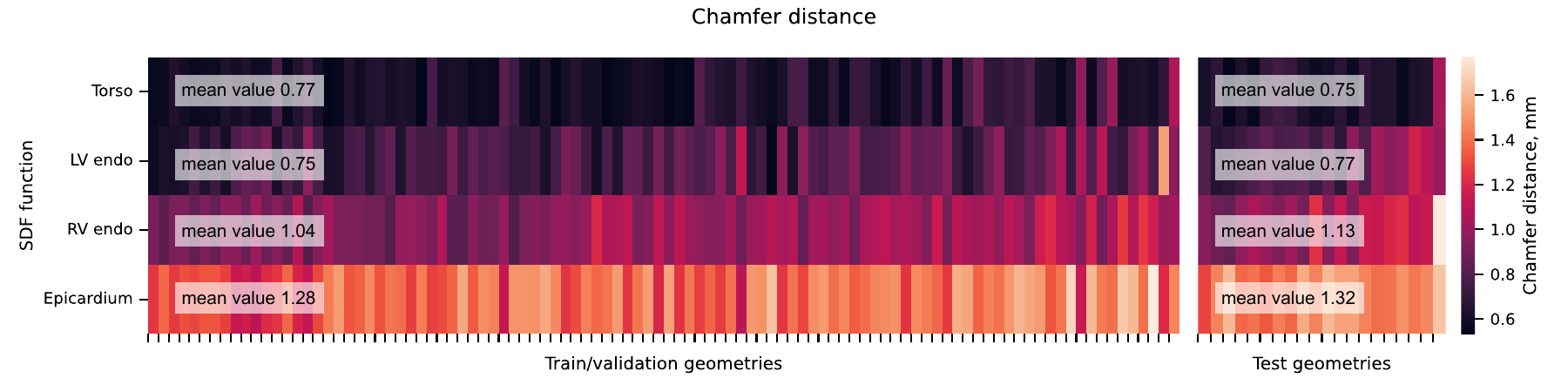}
    \caption{Chamfer distances (in mm) for the four SDF surfaces-torso, LV endocardium, RV endocardium, and epicardium, computed 80/20 training/validation geometries (left panel) and 10 test geometries (right panel). The color scale indicates the Chamfer distance value (in mm), with lighter colors corresponding to larger errors. For clarity, both the SDF surfaces and the geometry sets are sorted by increasing error.}
    \label{fig:sdf_heatmap}
\end{figure}

A heatmap of the reconstruction errors is shown in Fig.~\ref{fig:sdf_heatmap}.
Errors were consistently low across the torso and ventricular surfaces, indicating that the learned latent codes preserve the geometric detail required by the downstream lead-field surrogate. Test errors were comparable to training and validation errors for all surfaces, indicating good generalization.

Slightly larger errors were observed for the RV endocardium and the epicardium, likely due to their higher geometric complexity compared with the LV endocardium and the torso. The maximum error remained below \SI{1.6}{\mm}, while the mean error ranged between \SI{0.75}{\mm} and \SI{1.28}{\mm} across surfaces.

\subsection{Lead-field neural surrogate performance}

We report quantitative and qualitative results for the lead-field gradient prediction network, as well as for the ECG signals obtained by inserting the predicted $\nabla Z$ into Eq.~\eqref{eq:ECG}. 
Unless otherwise stated, FEM-based lead-field gradients are used as the reference solution.

For each of the 10 joint geometries from the test sample, SDF functions were assigned and test latent codes were calculated by solving the inference problem in Eq.~\eqref{eq:inference}.

\subsubsection{Qualitative comparison of gradient fields}

\begin{figure}[htbp]
    \centering
    \includegraphics[width=0.8\linewidth]{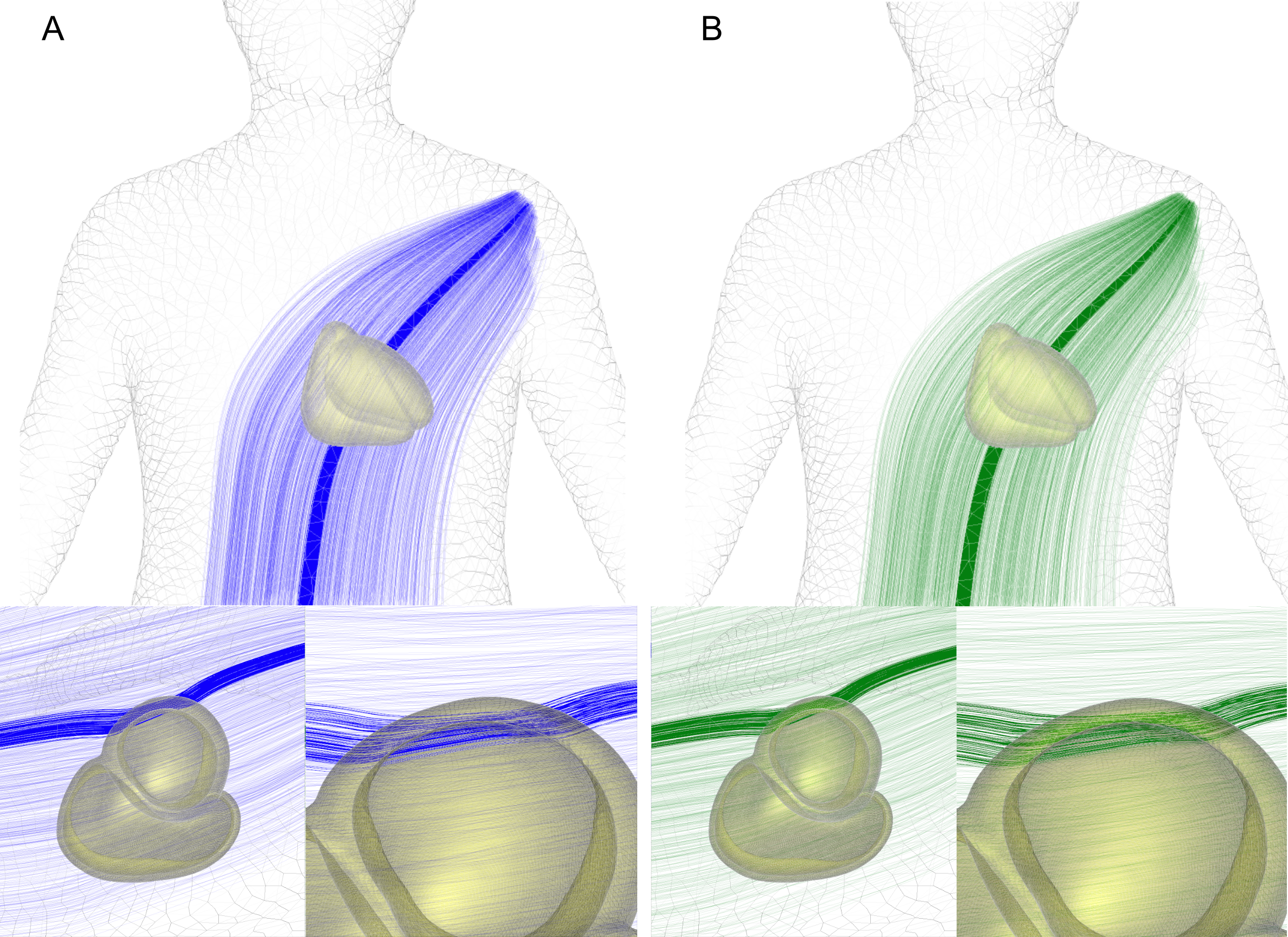}
    \caption{
    Streamline visualization of the lead-field gradient for a unipolar lead (left shoulder). 
    (A) FEM-based lead-field gradient. 
    (B) Predicted lead-field gradient. 
    Smaller sub-panels show close-up views of the heart and the LV free wall. 
    Streamlines intersecting the LV free wall are highlighted. 
    The bend in the streamlines at the heart-torso interface (highlighted in yellow) reflects the change in conductivity across tissues.
    }
    \label{fig:streamlines_compare}
\end{figure}

The surrogate accurately reproduces the FEM reference $\nabla Z$ across the torso volume. 
Fig.~\ref{fig:streamlines_compare} shows streamlines of the lead-field gradient (i.e., curves whose tangents coincide with the direction of $\nabla Z$) for a representative test geometry. 
Panel~A displays the FEM solution (blue), and Panel~B shows the surrogate prediction (green).

Visually, the predicted gradients closely match the ground truth in both direction and curvature. 
In particular, both solutions exhibit a clear change in direction at the heart-torso interface, reflecting the discontinuity in conductivity across tissues. 
This agreement indicates that the surrogate captures the dominant geometric and boundary-condition effects governing the lead field.

\subsubsection{Point-wise error distributions}

\begin{figure}[htbp]
    \centering
    \includegraphics[width=0.8\linewidth]{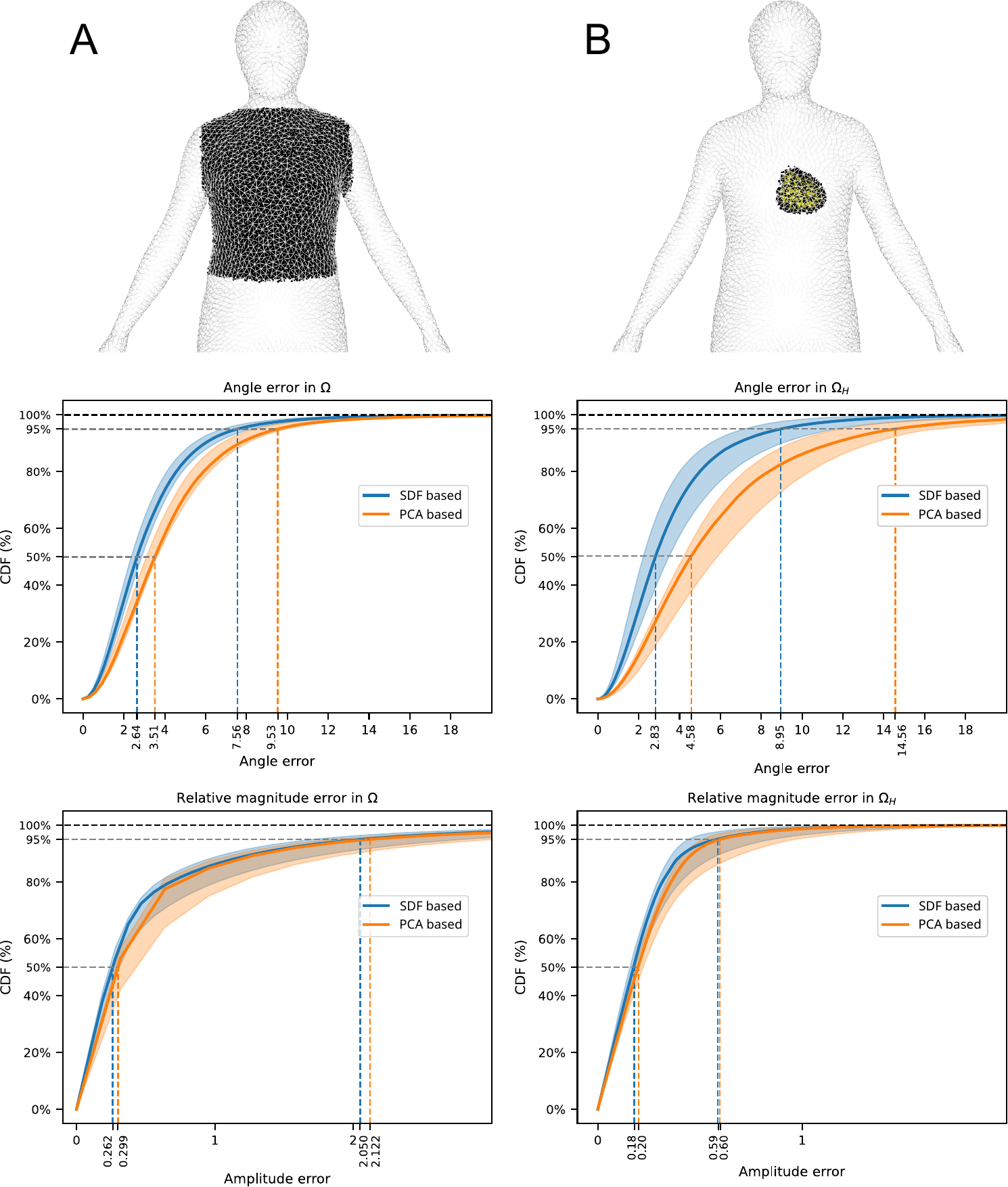}
    \caption{
    Cumulative distribution functions (CDFs) of prediction errors for $\nabla Z$ over the test dataset. 
    Left column: errors over the full torso domain. 
    Right column: errors restricted to points inside the heart and within \SI{10}{\mm} of the heart surface. 
    Top row: angular error (degrees). 
    Bottom row: relative magnitude error. 
    Blue and orange curves correspond to DeepSDF-based and PCA-based models, respectively. 
    Solid lines indicate the mean CDF across 10 test patients; shaded regions show the range (minimum--maximum). 
    Dashed vertical lines mark the median and $95^\text{th}$ percentile error values.
    }
    \label{fig:angular_cdf}
\end{figure}

Fig.~\ref{fig:angular_cdf} shows cumulative distribution functions (CDFs) of the angular error and relative magnitude error over the test cohort. 
Two evaluation regions are considered: 
(A) the full torso domain (matching the training sampling distribution), and 
(B) a restricted region consisting of points inside the heart and within \SI{10}{\mm} of the heart surface.

The heart-focused evaluation is motivated by two considerations. 
First, the ECG depends only on $\nabla Z$ within the heart domain $\Omega_H$ (see Eq.~\eqref{eq:ECG}). 
Second, the heart-torso interface is the region where $\nabla Z$ changes direction most abruptly, and is therefore potentially most sensitive to prediction errors.

The dashed vertical lines indicate the median ($50\%$) and $95\%$ quantile. 
For example, in the full-domain setting, $50\%$ of all test points exhibit an angular error below $2.64^\circ$ for the DeepSDF-based model and below $3.51^\circ$ for the PCA-based model. 
Across all configurations, the DeepSDF-based model consistently achieves smaller angular errors, while magnitude errors are comparable between the two models. 
The similarity in magnitude errors suggests that both models accurately capture the overall scaling of the gradient field, with differences primarily arising in directional alignment.

\subsubsection{Electrode-wise error analysis}

\begin{figure}[htbp]
    \centering
    \includegraphics[width=\linewidth]{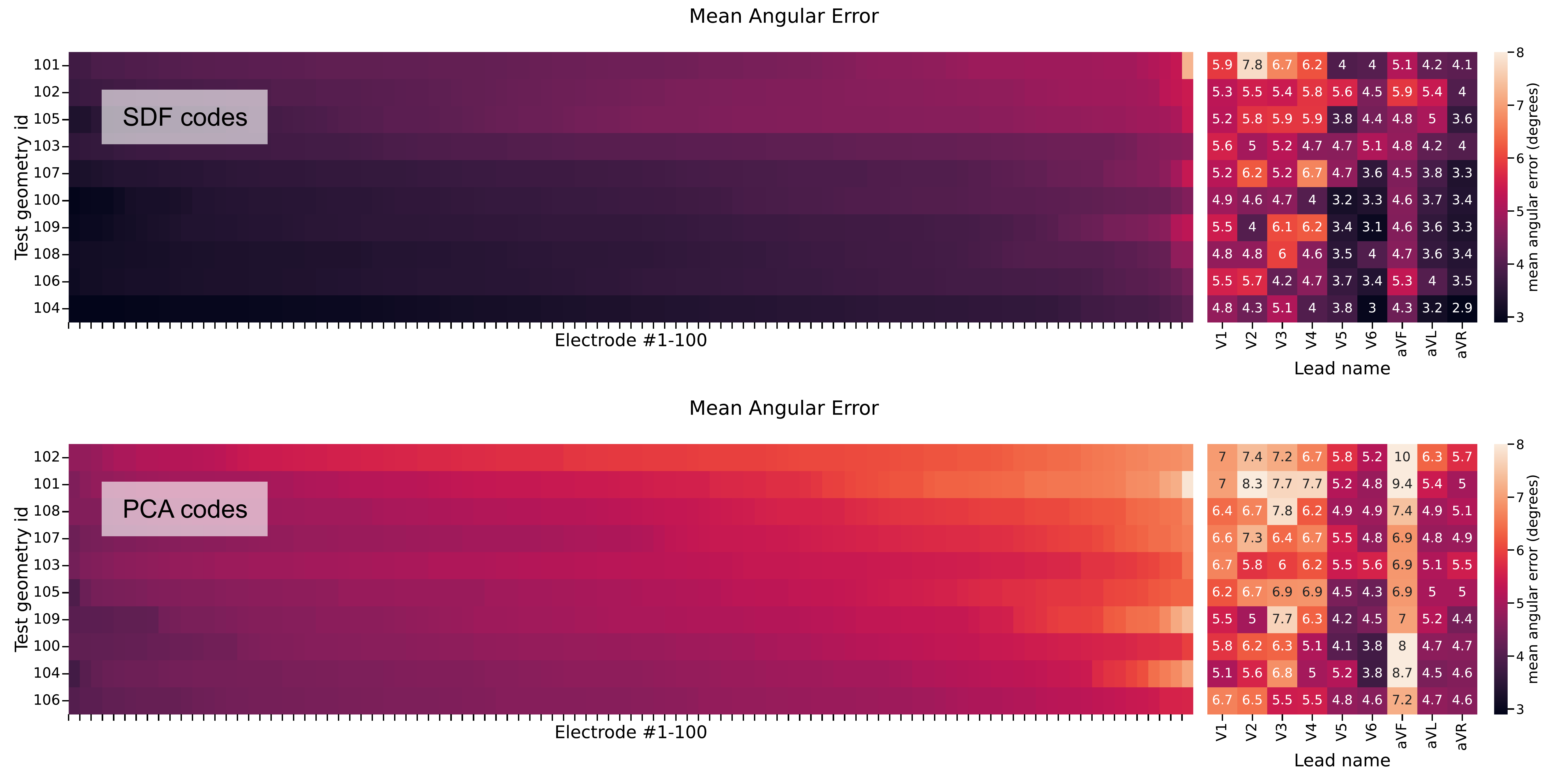}
    \caption{
    Average angular error (degrees) in predicting $\nabla Z$ using DeepSDF-based (top row) and PCA-based (bottom row) encodings. 
    Left column: errors across 100 uniformly distributed unipolar electrodes. 
    Right column: errors across the 9 independent electrodes of the standard 12-lead ECG. 
    Electrodes and patients are sorted by increasing error for visualization clarity. 
    }
    \label{fig:errors_heatmap}
\end{figure}

Fig.~\ref{fig:errors_heatmap} reports electrode-wise average angular errors for both encoding strategies, evaluated on 100 uniformly distributed unipolar electrodes and on the 9 independent electrodes of the standard 12-lead ECG.

The largest errors are observed for the aVF electrode and for precordial leads V1-V3. 
For precordial electrodes, this can be attributed to their proximity to the heart surface and thus to the heart-torso interface, where the gradient direction varies rapidly and is more challenging to approximate. 
In these regions, the gradient field exhibits higher curvature and stronger spatial heterogeneity, which amplifies directional errors.

The spatial distribution of angular errors, averaged over 10 test geometries and interpolated onto the torso surface, is shown in Fig.~\ref{fig:error_distribution}A. 
High-error regions (red) are primarily localized in the central anterior torso, consistent with the proximity to the heart.

\subsubsection{Computational efficiency}

We compared the computational cost of the proposed surrogate with the FEM-based lead-field computation. 
On a CPU implementation, solving the forward problem with FEM required approximately \SI{6}{\second} per lead, excluding mesh generation. Mesh construction required an additional \SI{22}{\second} per geometry.

In contrast, evaluation of the trained lead-field surrogate required approximately \SI{250}{\ms} per lead on a CPU (batch size = 1). 
Thus, even without GPU acceleration or batching, the surrogate provides a 24-fold speedup per lead.

\subsubsection{Impact on ECG signals}

\begin{figure}[htbp]
    \centering
    \includegraphics[width=\linewidth]{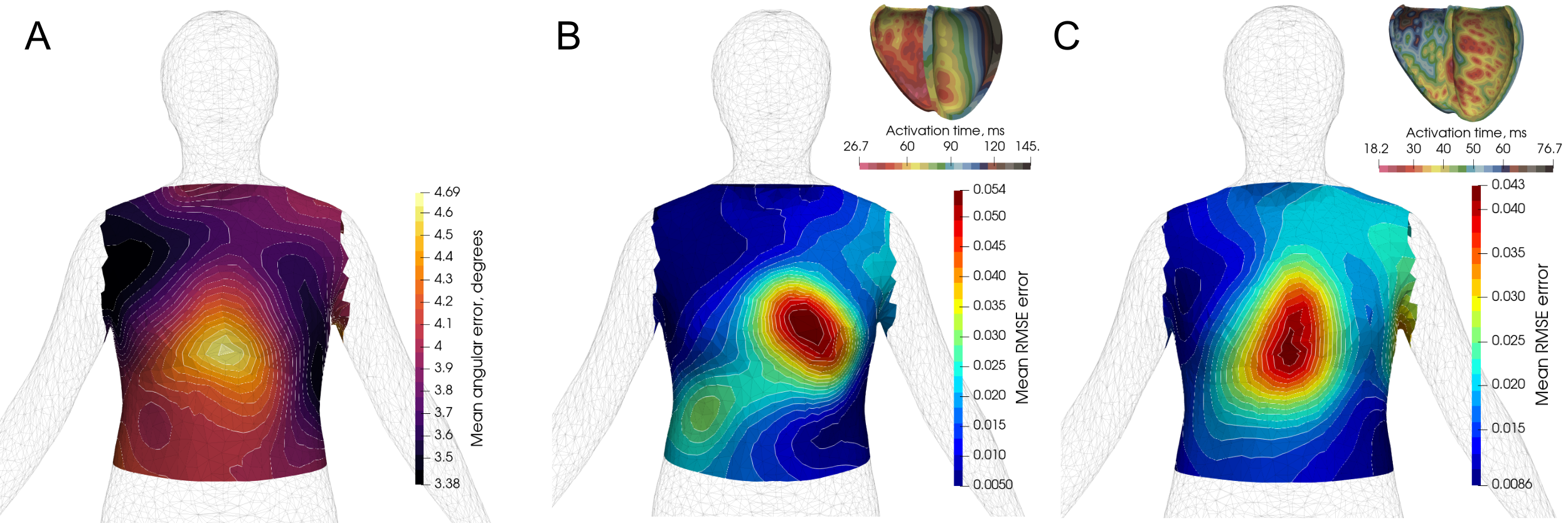}
    \caption{
    Spatial distribution of errors projected onto the torso surface and averaged over 10 test cases. 
    (A) Mean angular error (degrees) between DeepSDF-based predicted and FEM-based $\nabla Z$, computed over 100 electrodes. 
    (B--C) Mean RMSE of unipolar ECG signals for LBBB (B) and sinus rhythm (C), computed using FEM-based and predicted $\nabla Z$. 
    Errors are interpolated onto the torso surface; blue indicates smaller error. 
    Insets show representative ventricular activation maps used for ECG computation.
    }
    \label{fig:error_distribution}
\end{figure}

To assess practical relevance, we computed ECG signals using the predicted lead-field gradients. 
Fig.~\ref{fig:error_distribution}B-C shows the spatial distribution of the root mean squared error (RMSE) between body-surface potentials obtained with FEM-based and predicted $\nabla Z$ for two activation patterns: left bundle branch block (LBBB) and sinus rhythm. 
Regions of larger ECG error largely coincide with regions of larger angular gradient error, particularly in areas close to the heart.

\begin{figure}[htbp]
	\centering
	\includegraphics[width=\linewidth]{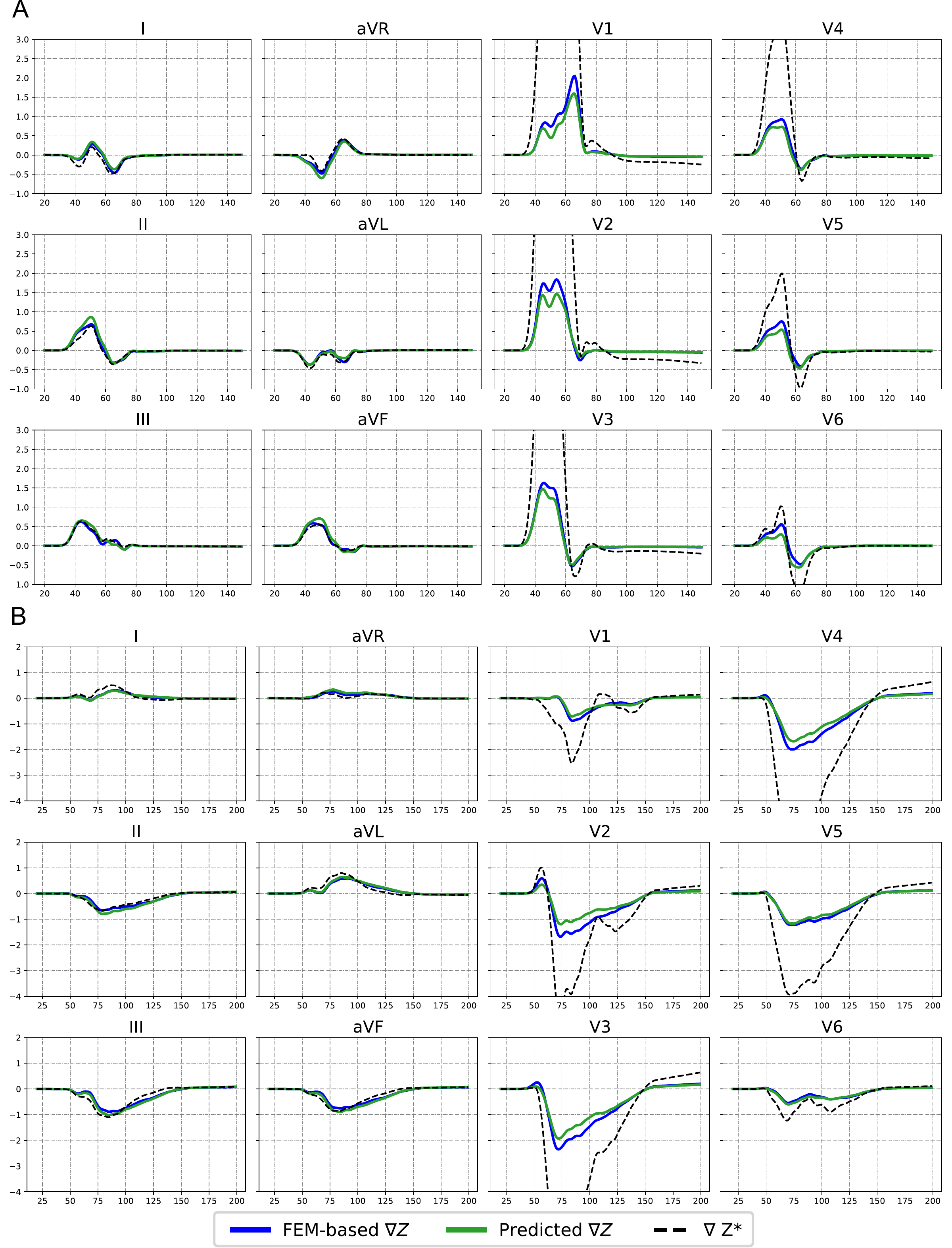}

    \caption{Forward ECG simulation for sinus rhythm (A) and LBBB (B) cases, computed using FEM-based (ground truth, blue), predicted $\nabla Z$ (green) and using the pseudo lead-field formulation $\nabla Z^*$.}
	\label{fig:results}
\end{figure}

Despite these localized discrepancies, the resulting ECG waveforms closely match the FEM-based signals (Fig.~\ref{fig:results}, left), indicating that the surrogate preserves clinically relevant waveform morphology.

Tab.~\ref{tab:quant_summary} summarizes quantitative performance on the test set. We report both field-level errors (accuracy of $\nabla Z$) and ECG-level errors. We observe that in general the DeepSDF-based approach is always performing better than the PCA-based encoding.

\begin{table}[htbp]
    \centering
    \begin{tabular}{llll}
        \toprule
        Geometry encoding & Angular error & Angular error (ECG leads) & ECG rel.\ $\ell_2$ error \\
        \midrule
        PCA-based & \SI{5.22 \pm 0.61}{\degree}
                  & \SI{5.93 \pm 1.26}{\degree}
                  & \num{0.024 \pm 0.013} \\
        DeepSDF-based & \SI{3.89 \pm 0.51}{\degree}
                      & \SI{4.64 \pm 0.99}{\degree}
                      & \num{0.018 \pm 0.01} \\
        \bottomrule
    \end{tabular}
    \caption{
    Average angular error (degrees) between FEM-based and predicted lead-field gradients $\nabla Z$ for points within the heart, and corresponding ECG relative $\ell_2$ error. 
    The third column reports angular error restricted to precordial ECG leads. 
    Values are given as mean $\pm$ standard deviation across test geometries.
    }
    \label{tab:quant_summary}
\end{table}



\section{Discussion}


In this work, we proposed a shape-informed neural surrogate model for the lead-field gradient arising in the coupled heart-torso problem, designed as a drop-in replacement for the full-order model in forward ECG simulations. Importantly, we did not surrogate the mapping from the transmembrane potential $\Vm$ to the ECG signal, i.e., the solution operator of the pseudo-bidomain problem~\eqref{eq:pseudo}. Instead, we focused exclusively on approximating the lead-field gradient. This design choice offers several advantages. The ECG can be expressed explicitly via Eq.~\eqref{eq:ECG}, which defines a linear functional acting on $\Vm$. Once the lead field is available, ECG computation is computationally inexpensive and remains fully general with respect to the underlying transmembrane potential and the intracellular conductivity tensor~$\tGi$. 
The latter plays a crucial role in determining ECG amplitude and morphology~\cite{Pezzuto2017-yw}. 
By surrogating only the lead-field gradient, we preserve this generality while replacing only the geometrically dependent component of the model, namely the mapping that embeds torso conductivity, electrode configuration, and anatomical variability.

A key strength of the proposed framework is that it is fully shape-informed. Still, for a new patient, only limited geometric information is required: once a surface representation is available (e.g., from segmented imaging data or sparse point clouds), inference of the latent code can be performed efficiently using either PCA projection or an inexpensive DeepSDF inference step~\cite{verhulsdonk24a}.
Such a mechanism is particularly attractive in the solution of the inverse problem of electrocardiography~\cite{VicentePuig2026, molero2023robustness} and in time-dependent settings, where cardiac geometry may evolve over time (e.g., during cardiac cycles or remodeling). Tracking geometric changes and updating the latent code enables rapid evaluation of the resulting impact on the ECG without repeated full-order simulations~\cite{zappon2024integrated}.
The proposed framework is also particularly attractive in settings requiring repeated forward simulations with varying electrode configurations.  For example, in electro-anatomical mapping procedures during catheter ablation~\cite{verma2015approaches}, the recording electrode is continuously repositioned, potentially acquiring thousands of electrograms. In such scenarios, the FEM-based lead-field computation is impractical in real time, whereas surrogate inference enables near-interactive evaluation.
More broadly, the method is well suited for scenarios involving dense electrode arrays or iterative inverse procedures.

In constructing the surrogate, we adopted a segregated approach that decouples geometry encoding from neural field regression. As shown in recent related work~\cite{Carrara2026}, such a strategy is particularly suitable in data-scarce biomedical settings. The geometric encoder (either PCA-based or DeepSDF-based) serves as a generative shape model that provides a compact, low-dimensional representation of joint heart-torso anatomy. 
This representation simplifies the input design of the neural surrogate and enables data augmentation through sampling in the latent space. In turn, this improves robustness and generalization when only a limited number of high-fidelity simulations are available.

From a quantitative standpoint, the surrogate achieves mean angular errors for the DeepSDF-based (resp.~PCA-based) encoding below \SI{4}{\degree} (resp.~slightly above \SI{4}{\degree}) and relative ECG errors below \SI{2}{\percent} (resp.~\SI{2.5}{\percent}) on the test set.  While such errors would not be considered negligible from a purely numerical analysis perspective, they are small in the context of forward ECG modeling. 
As shown in Sec.~\ref{sec:metrics}, the ECG error can be bounded in terms of the $L^2$ error of $\nabla Z$, with a multiplicative constant depending only on the electrophysiological activation. 
Therefore, controlling the gradient error directly limits the induced ECG error.
Angular errors are localized primarily near the heart-torso interface and, despite these localized discrepancies, the resulting ECG waveforms preserve clinically relevant morphology and amplitude. 
Compared with the classical pseudo lead-field approximation, the proposed method achieves substantially lower angular and ECG errors while retaining fast inference time. 
Both PCA-based and DeepSDF-based encodings yield accurate lead-field surrogates; however, quantitative results (Table~\ref{tab:quant_summary}) indicate a consistent advantage of the DeepSDF representation.  Specifically, the DeepSDF-based model achieves a lower mean angular error ($\SI{3.89 \pm 0.51}{\degree}$ versus $\SI{5.22 \pm 0.61}{\degree}$ for PCA) and a lower ECG relative $\ell_2$ error ($1.8\%$ versus $2.4\%$).  The improvement is particularly evident for precordial leads, where accurate representation of localized geometric detail near the heart-torso interface is critical.

In its current CPU-based implementation, the surrogate inference per lead is roughly $25\times$ faster than the FEM computation, even without GPU acceleration. 
This gap becomes more pronounced in configurations with many electrodes, such as BSPMs comprising hundreds of leads.

In the present study, myocardial fiber orientations were assigned using a rule-based method and kept fixed once generated from the anatomical geometry. Similarly, tissue conductivities were assumed fixed across geometries. While intracellular conductivity $\tGi$ influences ECG amplitude through Eq.~\eqref{eq:ECG}, its impact on the lead-field solution itself is expected to be limited, as the lead field primarily depends on torso geometry and electrode configuration. 
Nevertheless, extending the surrogate to include additional input parameters, such as conductivity variations or fiber anisotropy, is conceptually straightforward: these parameters could be appended to the latent code.  Such extensions would, however, increase the dimensionality of the input space and the associated increased requirement of training data.
Similarly, anatomical structures such as lungs and blood cavities introduce additional conductivity heterogeneities, and could be incorporated into the geometric encoding and surrogate framework.

Beyond the specific application to forward ECG simulation, this work contributes to a broader class of geometry-conditioned surrogate models for PDE operators~\cite{CATALANI2026, serrano2024aroma}. In line with recent developments in shape-informed surrogate modeling for cardiac mechanics~\cite{Carrara2026}, we demonstrate that the action of a parameter-dependent solution operator can be efficiently approximated by decoupling geometric encoding from field approximation. 
The idea of approximating geometry-dependent Green functions or solution kernels extends naturally beyond cardiac electrophysiology. 
In the present case, the lead-field function plays the role of a Green-type kernel for the pseudo-bidomain formulation, mapping cardiac source terms to body-surface potentials. 
Learning this operator in a geometry-aware manner enables efficient reuse across different activation patterns and electrophysiological states, without retraining or restricting the space of admissible transmembrane potentials.
Lead fields also arise in electroencephalography (EEG) and other bioelectric inverse problems, where accurate yet computationally efficient forward models are essential. 
More generally, many elliptic and parabolic PDEs admit integral representations in terms of Green's functions whose structure depends strongly on domain geometry. 
The proposed framework suggests a pathway toward learning such geometry-dependent kernels in a data-efficient manner.

\section*{Acknowledgements}

This work has been supported by the project PRIN2022 (MUR, Italy, 2023-2025, no.~P2022N5ZNP) ``SIDDMs: shape-informed data-driven models for parametrized PDEs, with application to computational cardiology'', funded by the European Union (Next Generation EU, Mission 4 Component 2).
F.B., S.~Pagani and F.R.\ acknowledge the grant Dipartimento di Eccellenza 2023-2027, funded by MUR, Italy.
F.B., S.Pagani, S.Pezzuto and F.R.\ are members of GNCS, ``Gruppo Nazionale per il Calcolo Scientifico'' (National Group for Scientific Computing) of INdAM (Istituto Nazionale di Alta Matematica).
F.B.\ acknowledges the ``INdAM - GNCS Project'', codice CUP E53C24001950001.
S.Pezzuto acknowledges the support of the CSCS-Swiss National Supercomputing Centre project no.~lp100 and the SNSF-FWF project ``CardioTwin'' (no.~214817).

\printbibliography
\end{document}